\batchmode
\makeatletter
\def\input@path{{D:/Dropbox/thesis/me//}}
\makeatother
\documentclass[english]{article}
\usepackage[T1]{fontenc}
\usepackage[latin9]{inputenc}
\usepackage{amsmath}

\makeatletter
\title{On the Equivalence of Factorized Information Criterion Regularization and the Chinese Restaurant Process Prior}

\author{Shaohua Li \\
  Nanyang Technological University, Singapore \\
  {\tt shaohua@gmail.com} \\ }

\makeatother

\usepackage{babel}
\begin{document}
\maketitle
\begin{abstract}
Factorized Information Criterion (FIC) is a recently developed information
criterion, based on which a novel model selection methodology, namely
Factorized Asymptotic Bayesian (FAB) Inference, has been developed
and successfully applied to various hierarchical Bayesian models.
The Dirichlet Process (DP) prior, and one of its well known representations,
the Chinese Restaurant Process (CRP), derive another line of model
selection methods. FIC can be viewed as a prior distribution over
the latent variable configurations. Under this view, we prove that
when the parameter dimensionality $D_{c}=2$, FIC is equivalent to
CRP. We argue that when $D_{c}>2$, FIC avoids an inherent problem
of DP/CRP, i.e. the data likelihood will dominate the impact of the
prior, and thus the model selection capability will weaken as $D_{c}$
increases. However, FIC overestimates the data likelihood. As a result,
FIC may be overly biased towards models with less components. We propose
a natural generalization of FIC, which finds a middle ground between
CRP and FIC, and may yield more accurate model selection results than
FIC. 
\end{abstract}

\section{Equivalence of FIC and CRP when $D_{c}=2$}

Suppose there are a sequence of $1$-of-$K$ latent coding variables
$\boldsymbol{Z}=\boldsymbol{z}_{1},\cdots,\boldsymbol{z}_{N}$. For
any $k$, let $n_{k}=\sum_{i=1}^{N}z_{ik}$. Then $\boldsymbol{Z}$
corresponds to a partition of $N$ numbers into $K$ sets $S_{1},\cdots,S_{K}$,
where $||S_{k}||=n_{k}$. This partition is denoted as $\boldsymbol{B}=(S_{1},\cdots,S_{K})$.
The correspondence between $\boldsymbol{Z}$ and the partition $\boldsymbol{B}$
is referred to as $\boldsymbol{Z}$ \textit{maps to }$\boldsymbol{B}$,
denoted as $\boldsymbol{Z}\mapsto\boldsymbol{B}$.

A Chinese restaurant process \cite{infgaussian,nonparatut} assigns
to this sequence a prior probability 
\[
P_{\textrm{CRP}}(\boldsymbol{Z})=\frac{\prod_{k}(n_{k}-1)!}{N!K!}.
\]

There are $\frac{N!}{\prod n_{k}!}$ configurations of $\boldsymbol{Z}$
mapping to the same $\boldsymbol{B}$. These configurations of $\boldsymbol{Z}$
form an equivalence class $\{\boldsymbol{Z}|\boldsymbol{Z}\mapsto\boldsymbol{B}\}$.
When it is clear from context, we also denote $\boldsymbol{B}=\{\boldsymbol{Z}|\boldsymbol{Z}\mapsto\boldsymbol{B}\}$.
The probability of this equivalence class is: 
\begin{equation}
P_{\textrm{CRP}}(\boldsymbol{B})=\frac{N!}{\prod n_{k}!}P(\boldsymbol{Z}_{0})=\frac{1}{K!\prod_{n_{k}>0}n_{k}},\label{eq:P_CRP}
\end{equation}
where $\boldsymbol{Z}_{0}$ is any configuration that maps to $\boldsymbol{B}$.

Note that $K$ is a free parameter. Fixing $K$ to a particular value,
we obtain a distribution of $\boldsymbol{B}$ conditioned on $K$:
\[
P_{\textrm{CRP}}(\boldsymbol{B}|K)=\frac{1}{\mathcal{Z}_{K}}\frac{1}{\prod_{n_{k}>0}n_{k}},
\]
where $\mathcal{Z}_{K}=\sum_{\begin{subarray}{c}
\forall n_{k}\ge0,\\
\sum n_{k}=N
\end{subarray}}\frac{1}{\prod_{n_{k}>0}n_{k}}$ is the normalizing constant.

When $D_{c}=2$, the FIC regularization term in \cite[eq.9]{fabmm}
is 
\begin{equation}
P_{\textrm{FIC}}(\boldsymbol{Z}|K)\sim\frac{1}{\prod_{n_{k}>0}(\sum z_{ik})}=\frac{1}{\prod_{n_{k}>0}n_{k}}.\label{eq:P_FIC}
\end{equation}

By comparing (\ref{eq:P_CRP}) and (\ref{eq:P_FIC}), one can see
that this regularizer term is equivalent to the CRP prior over the
equivalence class when the model parameter dimensionality $D_{c}=2$.

\section{Stronger Model Selection of FIC when $D_{c}>2$}

In higher dimensionality of $D_{c}$, $P_{\textrm{FIC}}(\boldsymbol{Z}|K)\sim\frac{1}{\prod_{n_{k}>0}n_{k}^{D_{c}/2}}$,
i.e. the FIC regularizer becomes sharper and more biased among different
configurations of $\boldsymbol{Z}$. To analyze the significance of
the exponent $D_{c}/2$, suppose we use a prior $p(\boldsymbol{Z})$
of $\boldsymbol{Z}$ in a model, where the data likelihood is given
by $p(\boldsymbol{X}|\boldsymbol{Z},\boldsymbol{\theta})$. The posterior
of $\boldsymbol{Z}$ is proportional to $p(\boldsymbol{X}|\boldsymbol{Z},\boldsymbol{\theta})p(\boldsymbol{Z})$. 

We first suppose the configuration of $\boldsymbol{Z}$ is known,
and consider a Laplace approximation of $p(\boldsymbol{X}|\boldsymbol{Z},\boldsymbol{\theta})$
w.r.t. $\boldsymbol{\theta}$. When $D_{c}$ is larger, $p(\boldsymbol{X}|\boldsymbol{Z},\boldsymbol{\theta})$
decreases more quickly when $\boldsymbol{\theta}$ deviates from the
ML estimator (MLE) $\bar{\boldsymbol{\theta}}$. That is, in $p(\boldsymbol{X}|\boldsymbol{Z},\boldsymbol{\theta})p(\boldsymbol{Z})$,
if $D_{c}$ is large enough, the effect of $p(\boldsymbol{Z})$ will
be dominated by $p(\boldsymbol{X}|\boldsymbol{Z},\boldsymbol{\theta})$,
if the prior $p(\boldsymbol{Z})$ does not change with $D_{c}$. In
other words, the regularization brought about by the prior will weaken
as $D_{c}$ increases. The CRP prior is covered by this analysis.
In contrast, as the FIC regularizer becomes sharper as $D_{c}$ increases,
the domination of $p(\boldsymbol{X}|\boldsymbol{Z},\boldsymbol{\theta})$
over $p(\boldsymbol{Z})$ will not happen with the increase of dimensionality
of $\boldsymbol{\theta}$. This suggests that FIC will have a stronger
model selection effect and tend to be more ``parsimonious'' than
CRP, when the parameter dimensionality is high.

\section{Possible Limitations and Generalizations of FIC}

The analysis in Section 2 is based on Laplace approximation, in which
the approximating Gaussian is only accurate around a small area around
the MLE $\bar{\boldsymbol{\theta}}$ of $p(\boldsymbol{X}|\boldsymbol{Z},\boldsymbol{\theta})$.
It might be the case that the estimated marginal probability is greater
than the actual probability. Actually during the derivation of Factorized
Asymptotic Bayesian inference, \cite{fabmm} assumes that the MLE
$\bar{\boldsymbol{\theta}}$ of $\sum_{\boldsymbol{Z}}p(\boldsymbol{X}|\boldsymbol{Z},\boldsymbol{\theta})p(\boldsymbol{Z})$
is also the MLE of $p(\boldsymbol{X}|\boldsymbol{Z},\boldsymbol{\theta})$
for each particular $\boldsymbol{Z}$. Therefore the estimated marginal
probability would be greater than the actual marginal probability.

In this regard, we could extend FIC to a Generalized FIC (GFIC), which
is milder thanks to a smaller exponent, i.e. $P_{\textrm{GFIC}}(\boldsymbol{Z}|K)\sim\frac{1}{\prod_{n_{k}>0}n_{k}^{d}}$,
where $1<d<D_{c}/2$.

\bibliographystyle{plain}
\bibliography{3D__Dropbox_thesis_me_fabfhmm}

\end{document}